\documentclass[twoside,11pt]{article}
\usepackage[rightcaption]{sidecap}
\usepackage{blindtext}
\usepackage{amsmath}
\usepackage{booktabs}
\usepackage{multirow}
\usepackage[abbrvbib, preprint]{jmlr2e}

\usepackage{lastpage}

\firstpageno{1}

\begin{document}

\title{Lighting-Aware Representation Learning under Controllable Lighting Variation}

\author{\name Lizhen Zhu \email ljz5180@psu.edu\\
       \name Charantej Reddy Pochimireddy \email charantejreddy.p@gmail.com \\
       \name James Z. Wang \email jwang@ist.psu.edu\\
       \addr Department of Informatics and Intelligent Systems\\
       College of Information Sciences and Technology\\
       The Pennsylvania State University\\
       University Park, PA, 16802, USA
       \AND
       \name Brad Wyble \email bpw10@psu.edu \\
       \addr Department of Psychology \\
       College of the Liberal Arts\\
       The Pennsylvania State University\\
       University Park, PA, 16802, USA}
       
\editor{My editor}

\maketitle

\begin{abstract}
Variations in illumination remain a major challenge for visual representation learning, as they induce substantial appearance changes both across and within environments. While existing approaches typically address this issue through data augmentations that encourage models to become invariant to lighting changes, such strategies do not explicitly model lighting information during learning.
Inspired by theories of human vision, we propose a lighting-aware representation learning framework that incorporates illumination variation as an explicit training signal rather than a nuisance factor to be suppressed. Our method extends contrastive learning by introducing an auxiliary objective that captures illumination-dependent variation in rendered scenes, enabling the model to jointly learn representations that preserve semantic consistency while remaining sensitive to lighting-dependent visual structure.
We evaluate the proposed model on image classification and object detection tasks across the ImageNet, ExDark, and PASCAL VOC benchmarks. Results demonstrate that the proposed lighting-aware training consistently improves downstream performance over standard contrastive learning baselines, while maintaining the same architecture and training budget. Furthermore, our approach shows promising performance in supervised learning frameworks and under settings involving simpler lighting variation, suggesting broad applicability beyond complex illumination scenarios. These results indicate its potential to enhance model robustness and adaptability in complex visual environments as well as in more conventional image processing tasks.
\end{abstract}


\section{Introduction}
\begin{figure}[h!]
    \centering
    \includegraphics[width=0.9\linewidth,
    trim=0 0 0 0,clip]{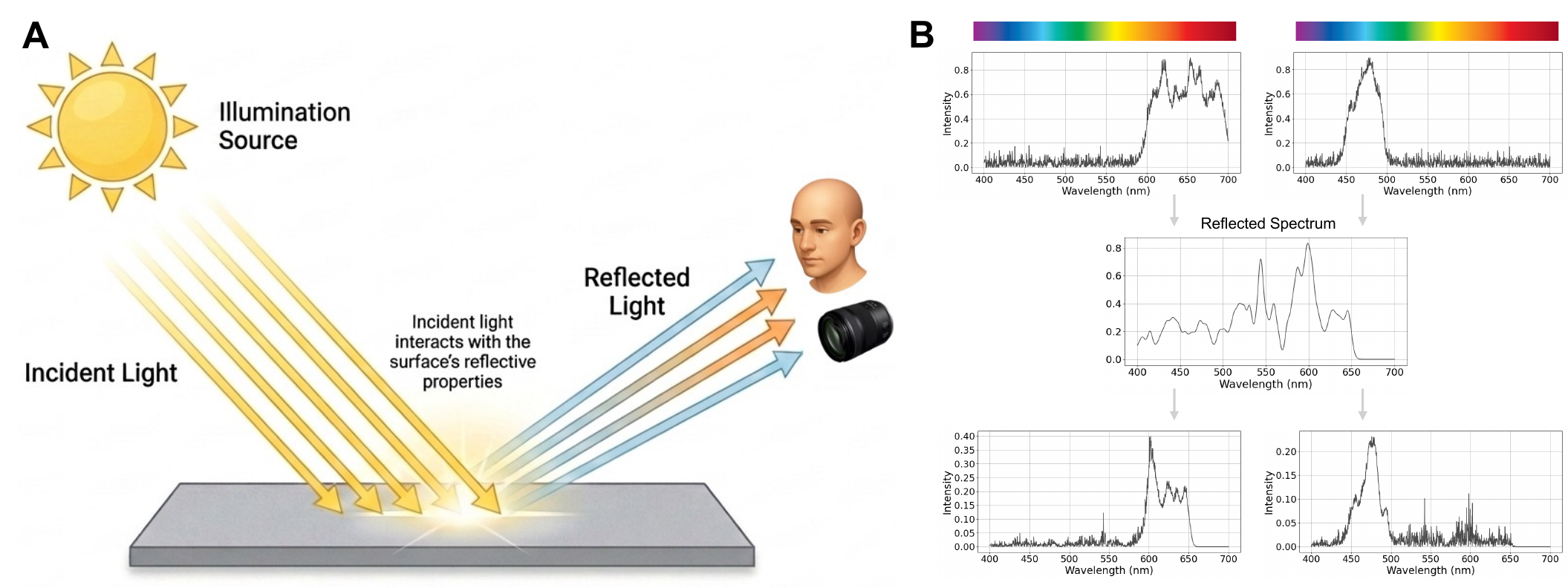}
    \caption{
    Illumination as a source of visual variation.
    (A) Scene appearance depends on illumination and surface reflectance.
    (B) Different illuminants produce different reflected spectra.
    }
    \label{fig:motivation}
\end{figure}

A fundamental challenge in computer vision is the reliable recognition of objects or scenes under varying illumination conditions. Both the human visual system and artificial image sensors detect light reflected from surfaces, where this reflectance arises from the interaction between the incident illumination and the intrinsic reflective properties of the surface (Figure~\ref{fig:motivation}A). Therefore, changes in the spectral distribution of the illuminating photons alter the observed appearance of an object, introducing substantial appearance variation in visual perception and image analysis (Figure~\ref{fig:motivation}B). Formally, an image may be viewed as an observation generated under specific scene and lighting conditions. Variations in illumination therefore introduce controllable variation in the visual signal, often leading to distribution shifts that challenge the robustness and generalization of computer vision systems.

Illumination variation poses a major challenges for real-world computer vision applications; systems trained in one environment often fail to recognize the same objects in different environments, or even at different times of day, due to changes in sunlight or reflected ambient light~\citep{tung2019large, pilligua2025evaluating}. Existing approaches attempt to learn lighting-invariant visual representations through image augmentation with synthetic illumination transformations~\citep{8628742,xiao2024generalization,shorten2019survey}. While these augmentations are effective, they still fall short of the degree of lighting robustness exhibited by human vision. This limitation may partly stem from sensor constraints, as conventional digital imaging sensors lack the dynamic range and luminance adaptation capabilities of the human visual system~\citep{skorka2011toward}. The human eye and brain together constitute a highly adaptive perceptual system capable of compensating for substantial lighting variations. Although hardware-based solutions may help address this challenge, they are often impractical due to increased cost or the requirement for acquiring multiple bracketed exposures under different camera settings. Instead, we pursue computational mechanisms inspired by theoretical models of human vision.

There is extensive literature documenting the capabilities and likely mechanisms of \textit{perceptual constancy} in human vision~\citep{smithson2005sensory}, the ability to perceive objects as stable across a wide range of viewing conditions, including variation in luminance. A recurring theme in theories explaining human visual robustness to illumination variation is that the visual system estimates the illuminant within a scene and constructs an internal model of the lighting condition. This model is then used to compensate for illumination variation relative to a baseline scene representation. Notably, this strategy differs substantially from standard augmentation-based approaches in computer vision, which do not attempt to learn explicit representations of lighting. Instead, such methods train networks to suppress randomized variations in image properties associated with illumination, such as hue and contrast.

\begin{figure}[ht!]
    \centering
    \includegraphics[width=0.9\linewidth]{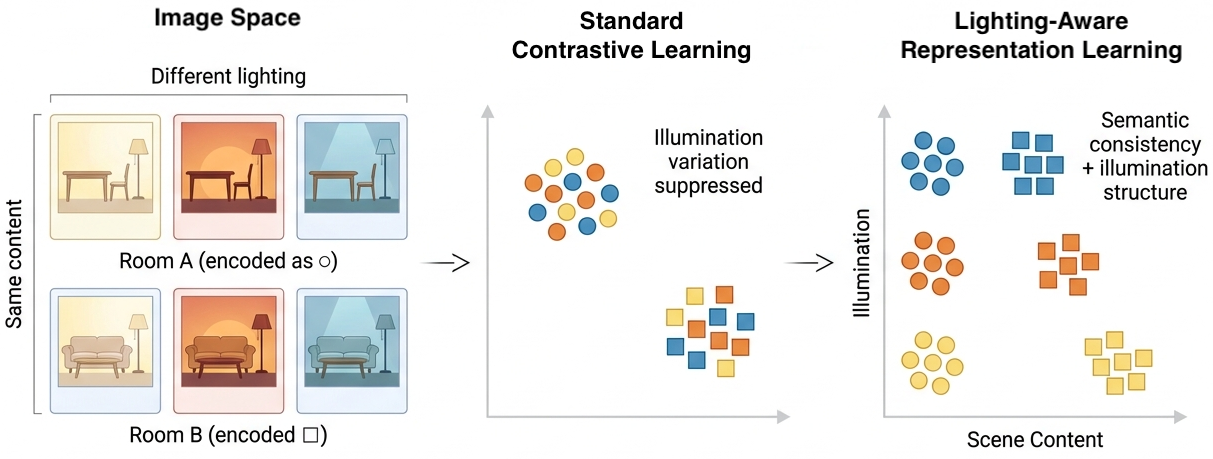}
    \caption{
    Illustrative Comparison of embeddings produced by standard contrastive learning, and the desired embedding produced by a lighting-aware framework.}
    \label{fig:representation}
\end{figure}

Motivated by this perspective, we ask whether explicitly modeling lighting can lead to more robust visual representations than treating it as noise. Figure~\ref{fig:representation} provides a simplified conceptual illustration of the desired organization of the learned representation. In this work, we propose a learning framework that incorporates illumination variations directly into the training objective. To enable {\it controlled} study of illumination effects, we use the House100KLighting dataset~\citep{zhu2024incorporating}, collected from a virtual ray-traced environment~\citep{gan2020threedworld}, which permits precise manipulation of lighting conditions while preserving underlying scene content. We extend contrastive learning by introducing an additional objective that encourages representations to capture lighting-dependent structure alongside semantic content similarity.

\begin{figure}[ht!]
    \centering
    \includegraphics[width=0.9\linewidth,
    trim={0 0cm 0 0cm},clip]{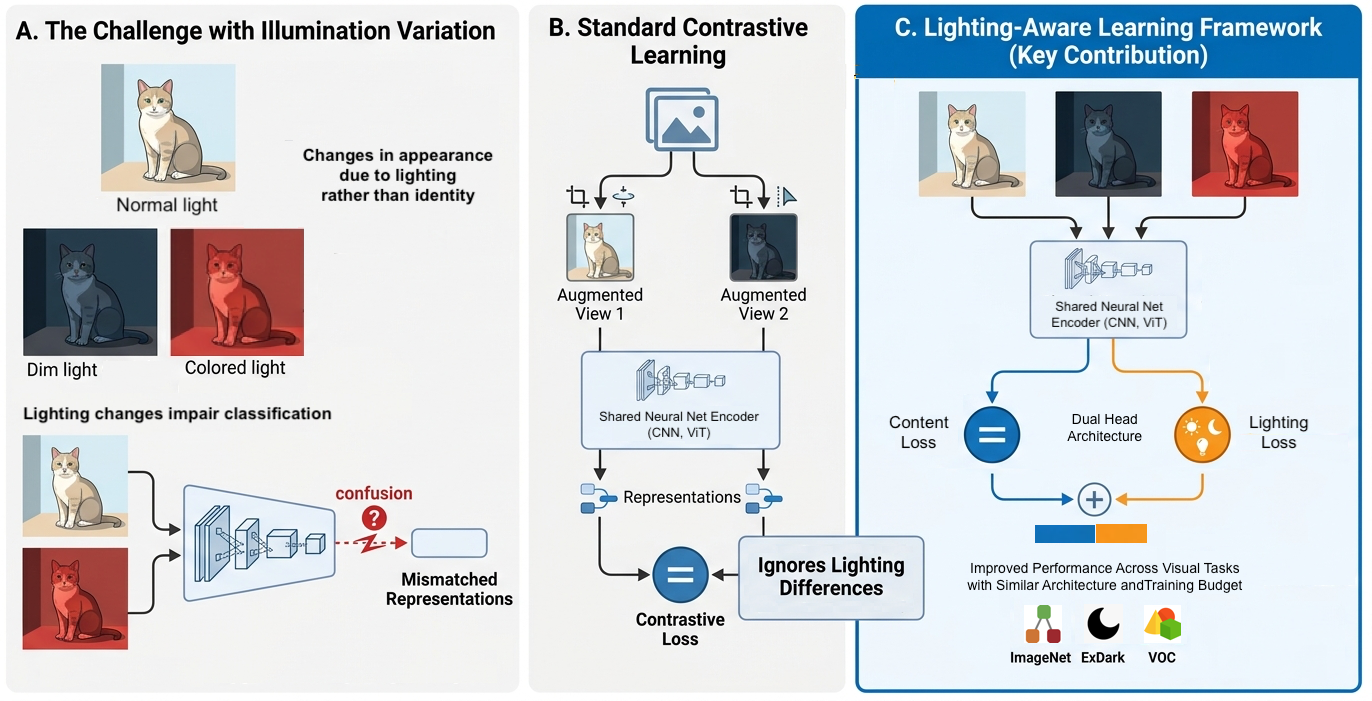}
    \caption{
    Overview of the proposed lighting-aware representation learning framework, which incorporates illumination as an explicit training signal to jointly learn semantic and lighting-aware representations.
    }
    \label{fig:teaser}
\end{figure}

Our approach jointly learns two complementary representations of visual data: one that organizes them according to illumination conditions and another that groups images according to semantic content (Figure~\ref{fig:teaser}). This design encourages the model to learn factor-aware representations, that support semantic consistency across lighting conditions while remaining sensitive to lighting-dependent visual structure when it provides informative environmental cues.

We evaluate the proposed method on visual recognition benchmarks, including image classification and object detection tasks. Across experiments, models trained with the proposed lighting-aware objectives achieve consistently higher predictive performance compared to counterparts that rely solely on standard contrastive learning or augmentation-based strategies despite using the same backbone and total number of training epochs.

This work investigates whether controllable environmental variation can be leveraged as an explicit source of supervision for representation learning rather than treated solely as nuisance variability. We use illumination as a case study because it produces substantial appearance variation while preserving scene semantics, and also because modern simulation environments allow lighting conditions to be manipulated precisely at scale. Our results suggest that explicitly modeling such factors yield representations that are more effective for visual classification than approaches that seek invariance through augmentation alone. More broadly, we view illumination as one example of a wider class of controllable factors that influence image formation. This perspective points toward a general framework for factor-aware representation learning that could be extended to variables such as scale, viewpoint, time-of-day, and weather, ultimately improving robustness and generalization in complex, real-world visual environments.

\section{Related Work}
\label{sec: related_work}
\subsection{Contrastive Learning}
Contrastive learning has recently gained significant attention for its ability to learn useful representations without requiring labeled data, offering a learning paradigm loosely analogous to aspects of early human perceptual development. Early methods, such as MoCo~\citep{he2020momentum} and SimCLR~\citep{chen2020simple}, encourage models to bring similar images (positives) closer in feature space while pushing dissimilar ones (negatives) farther apart, thereby allowing the extraction of meaningful features from unlabeled data. Later advancements, including BYOL~\citep{grill2020bootstrap} and SimSiam~\citep{chen2021exploring}, introduced frameworks that eliminate the need for negative samples, further broadening the applicability of contrastive learning. 

More recent work has explored incorporating auxiliary supervision or structured training objectives into contrastive learning to improve robustness and representation quality across different domains~\citep{li2021disentangled, ZHANG2025106781, gu2023cddsa}. Our work builds on the contrastive learning paradigm by incorporating explicit lighting supervision to encourage representations that remain robust across varying illumination conditions while preserving lighting-dependent visual structure.

\subsection{Lighting-Aware and Illumination-Robust Vision}
Existing approaches often address the illumination variation through image augmentation strategies that simulate lighting changes using transformations such as brightness, contrast, hue, or gamma adjustment~\citep{8628742,xiao2024generalization,shorten2019survey}. Although these methods improve robustness, they typically encourage invariance to illumination without explicitly incorporating lighting information into representation learning.

Several prior works have explored modeling illumination-related information in visual processing pipelines. Retinex-based methods~\citep{597272,6176791,wei2018deep} separately model illumination and reflectance information for image enhancement, while recent approaches leverage deep learning architectures for adaptive low-light processing~\citep{li2018lightennet,jiang2021enlightengan,cai2023retinexformer}. In parallel, simulation platforms and controllable rendering environments~\citep{gan2020threedworld} have enabled systematic studies of visual robustness under varying environmental conditions. Our work differs from these approaches by focusing on lighting-aware representation learning under controlled lighting variation within a contrastive learning framework.

\subsection{Low Light Image Enhancement}
Since computer vision systems are frequently deployed in real-world environments with variable illumination, low-light image enhancement remains an important challenge in this field. Traditional methods, such as the Retinex algorithm~\citep{597272,6176791}, which is grounded in theories of color constancy, and LIME~\citep{guo2016lime}, which estimates pixel-wise illumination, have been widely adopted for low-light image enhancement. Recent advancements leverage deep learning to achieve improved low-light enhancement. For instance, LightenNet~\citep{li2018lightennet} introduces a CNN-based approach for adaptive light enhancement.
Based on the Retinex theory, Wei et al. decomposes images into reflectance and illumination components to separately enhance lighting~\citep{wei2018deep}. Self-supervised approaches such as  EnlightenGAN~\citep{jiang2021enlightengan} have shown strong performance in unstructured environments, while more recent architectures, such as Retinexformer~\citep{cai2023retinexformer}, incorporate Transformer-based architecture within the Retinex framework to capture richer contextual information. Although our work does not explicitly address low-light enhancement, we posit that learning lighting-aware representations may improve visual understanding under low-light and other challenging illumination conditions.

\section{Method}
\label{sec: methodology}

\subsection{Background: Contrastive Learning}
We build on contrastive learning frameworks that learn representations by bringing positive pairs closer while pushing negative pairs farther apart. Specifically, we adopt MoCo V2~\citep{chen2020improved} and its extension ESS-MB~\citep{zhu2024incorporating} as baseline methods.

\subsubsection{MoCo V2}

MoCo V2 learns visual representations using a momentum encoder and a memory bank of negative samples. Given an input image $I$, two augmented views are encoded into query and key embeddings, $q_i$ and $k_i$, respectively. The contrastive objective encourages similarity between positive pairs while discriminating them from other samples in the memory bank:
\begin{equation}
 L_\text{Con.}=- \log\frac{\exp(\text{sim}(q_i, k_{i})/\tau)}{\sum_{b\in B}\exp(\text{sim}(q_i, k_b)/\tau)}\;.
\end{equation}
Here, $\text{sim}(\cdot,\cdot)$ denotes cosine similarity function, $\tau$ is a temperature parameter, and $B$ denotes the set of negative samples.

\subsubsection{ESS-MB}

ESS-MB extends the standard contrastive framework by defining positive pairs based on spatial proximity in a simulated 3D environment. Images captured within a small specified threshold of positional and orientational similarity are treated as positives, enabling the model to learn from nearby viewpoints of the same scene. The objective averages the contrastive loss over all positive pairs P(i) associated with the anchor image i:
\begin{equation}
\label{eq:multi}
 L_\text{Con.}=-\frac{1}{|P(i)|}\sum_{p\in P(i)} 
 \log\frac{\exp(\text{sim}(q_i, k_{p})/\tau)}{\sum_{b\in B}\exp(\text{sim}(q_i, k_b)/\tau)}\;.
\end{equation}
These formulations provide the foundation for the lighting-aware objectives introduced in the following section.

\begin{figure}[ht!]
\begin{center}
    \includegraphics[width=\linewidth]{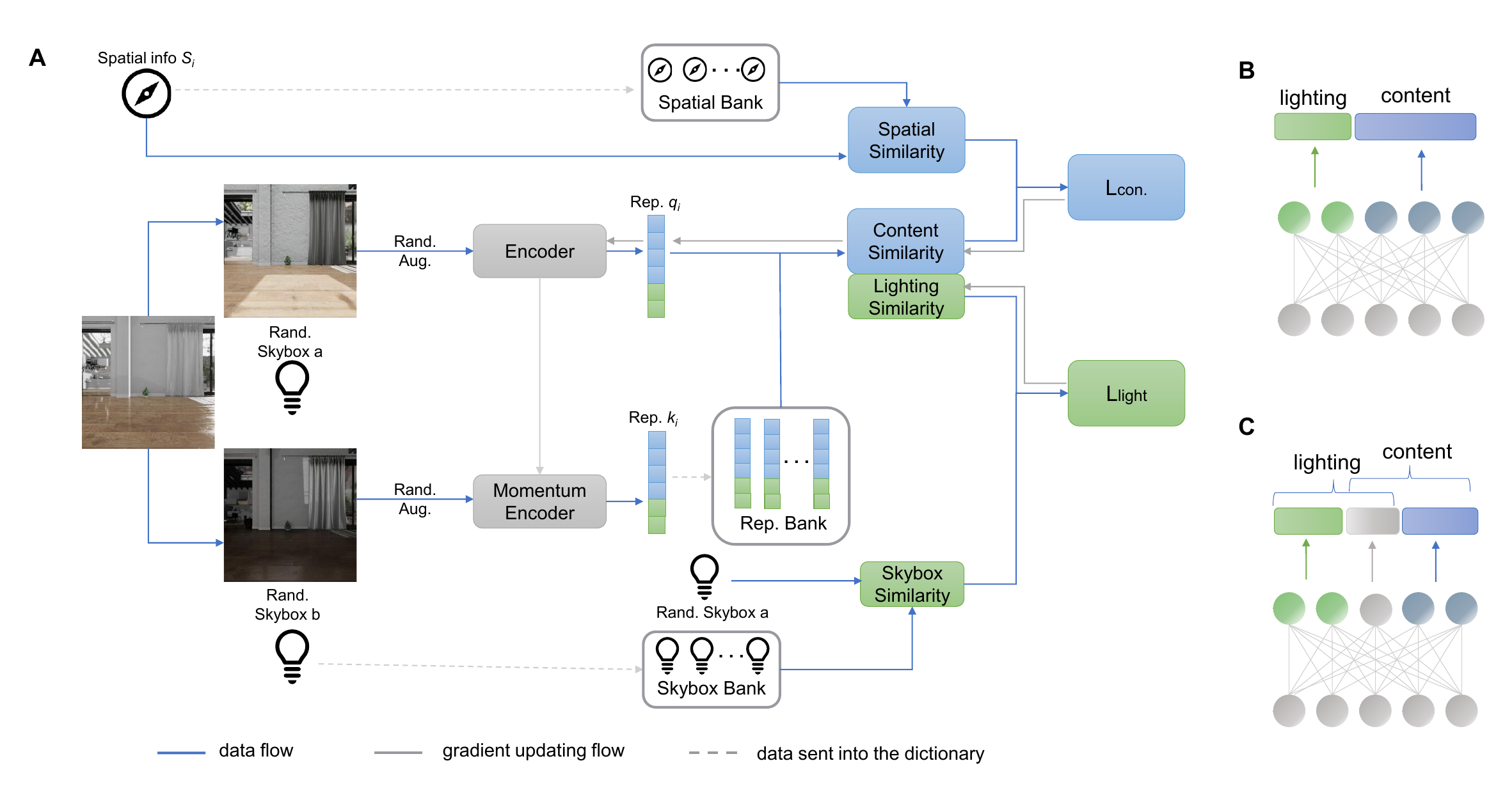}
\end{center}
\caption{Illustration of the proposed approach. (A) Pipeline of the proposed dual-head contrastive learning approach. (B) Division of the representation into lighting and content. (C) Division of the representation into lighting (green), joint (red), and content (blue) parts.} 
\label{fig:pipeline}
\end{figure}

\subsection{Lighting-Aware Dual-Head Framework}

We model each observed image $I$ as being generated from two underlying factors: the intrinsic scene content $C$ and the illumination condition $L$:
\begin{equation}
\label{eq:content_lighting}
I \sim p(I \mid C, L)\;.
\end{equation}
Here, $C$ represents scene-intrinsic information such as object geometry, layout, and semantic content, while $L$ characterizes illumination-related properties. Variations in $L$ introduce controllable appearance into the visual signal, motivating the learning of representations that remain consistent with respect to scene content while remaining sensitive to illumination conditions.

Building on this motivation, a dual-head training framework is designed to encourage complementary lighting-aware and content-aware representations. As shown in Figure~\ref{fig:pipeline} A, each scene is rendered under two lighting configurations, producing paired observations that share identical content but differ in illumination. The resulting images are further subjected to standard augmentations (e.g., random cropping and color perturbations), which are excluded from the lighting-specific loss function, and are then encoded by a ResNet-50 backbone within a modified contrastive learning framework.

The shared encoder branches after global average pooling into two heads: a contrastive head that enforces content consistency across varying lighting conditions, and a lighting head that predicts the lighting condition labels. The two heads differ only in their final layers, while all preceding layers are shared and jointly optimized under both objectives. This architecture introduces an inductive bias that encourages the organization of representations into illumination-related and content-specific components.

\subsection{Supervised Lighting Contrastive Head}
The lighting contrastive head aligns representations of images that share the same illumination configuration while separating those under different configurations. It takes the pooled features and passes them through a two-layer projection head, following standard contrastive learning practice. During training, this component performs contrastive learning by comparing the representations of the current image with those in the memory bank, under different illumination settings:
\begin{equation}
\label{eq:lightloss}
 L_\text{Light}=-\frac{1}{|L(i)|}\sum\limits_{l\in L(i)} \log\frac{\exp(\text{sim}(q^\prime_i, k^\prime_{l})/\tau)}{\sum_{b\in B}\exp(\text{sim}(q^\prime_i, k^\prime_b)/\tau)}\;,
\end{equation}
where $L(i)$ denotes the set of representations that share the same illumination condition as image $i$. $q^\prime_i$ and $k^\prime$ denote the lighting representations of image $i$ and the corresponding representation in the memory bank, respectively. 

\subsection{Dual-head Training}
During training, the dual-head structure encourages the network to differentiate between internal information that contributes to the lighting-aware representations and those representing the content. The overall loss function combines the lighting contrastive loss defined in Eq.~\eqref{eq:lightloss} and the standard contrastive loss from Eq.~\eqref{eq:multi}:
\begin{equation}
\label{eq:loss}
 L=L_\text{Con.}+ \alpha L_\text{Light}\;.
\end{equation}
Here, $\alpha$ is a weighting coefficient that balances the contributions of content and illumination contrastive training. Both illumination and content information contribute to the backpropagation process, leading to joint updates across the shared network layers prior to the final split into the two heads. 

\subsection{Overlapping Lighting and Content Representation}
\begin{figure}[ht!]
    \centering
    \includegraphics[width=0.9\linewidth]{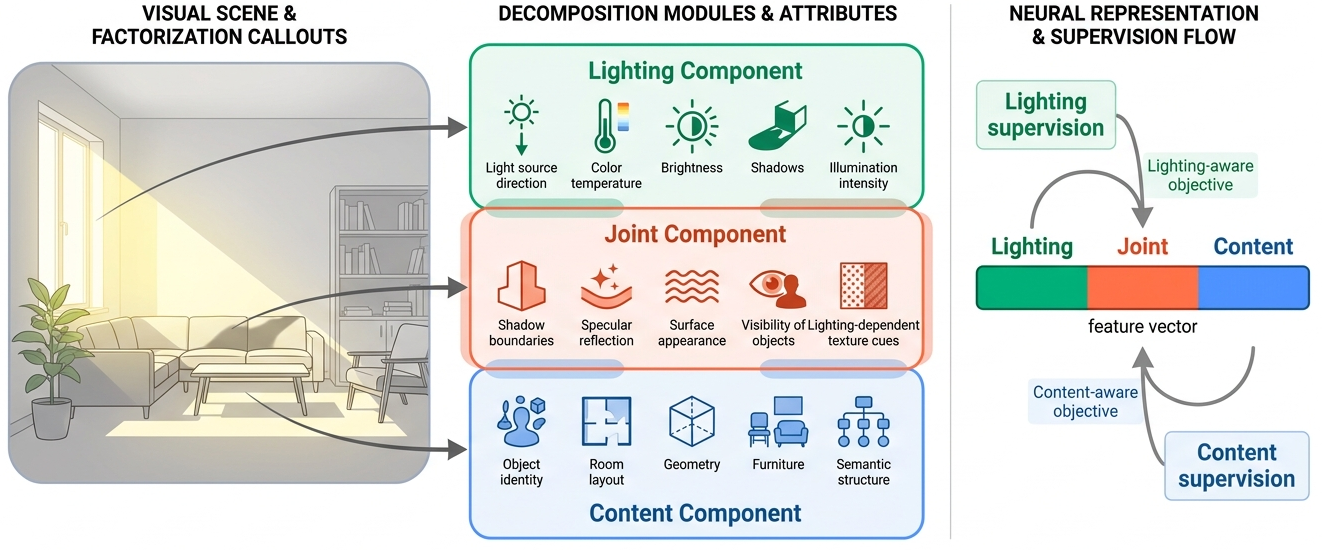}
    \caption{
    Some visual attributes are primarily determined by illumination (green),
    others by scene content (blue), while interaction effects such as shadows
    and reflections are captured by a shared joint representation (orange).
    }
    \label{fig:component}
\end{figure}

Naively enforcing a strict separation between lighting and content may impose an overly strong independence assumption, as real-world visual signals often exhibit statistical dependencies between these factors. To provide greater flexibility, we partition the learned representation into three components: lighting, content, and a joint component, as illustrated in Figure~\ref{fig:pipeline} C. Figure~\ref{fig:component} illustrates this factorization, highlighting that some visual information is primarily associated with illumination, some with scene content, and some with interactions between the two factors.

The lighting component captures features primarily determined by illumination conditions. The content component encodes scene-intrinsic information, such as object shapes, layout, and semantic identity, that is largely invariant to illumination. The joint component captures variations that cannot be cleanly attributed to either illumination or content alone, representing shared or interacting information between these factors. It is supervised by both lighting and content objectives, allowing the model to allocate representational capacity to subtle cross-factor interactions without enforcing strict independence.

To maintain a constant number of parameters, we keep the overall dimensionality of the representation fixed and allocate portions of it to the three components: lighting, joint, and content. We denote the allocation using a simple ratio notation $(L:J:C)$, where each term specifies the relative fraction of the projection head assigned to the corresponding component. For example, $(1:0:7)$ indicates that one-eighth of the representation is allocated to the illumination component, none to the joint component, and seven-eighths to the content component. This notation is used consistently in subsequent experiments to describe the relative composition of the representation in each configuration.

During training, the illumination-aware and joint components are optimized under the lighting contrastive objective, while the content and joint components are optimized under the content contrastive objective. This structured training scheme encourages the model to organize representations according to both semantic content and illumination conditions, while allowing shared components to capture interactions between these sources of variation.

\section{Experiments}
\label{sec: experiments}
We aim to address the following key research questions:
\begin{itemize}
\item What advantages do the representations learned through dual-head training provide, and what is the role of each component within the lighting-aware representation?
\item Can lighting conditions generated by ray tracing be effectively approximated by simpler image-level transformations?
\item Can this approach be extended to other training settings, such as supervised learning, while preserving robustness to lighting variation?
\end{itemize}

\subsection{Pretrain Datasets}
In the pretraining phase, we used two base datasets: House100K~\citep{zhu2024incorporating} and ImageNet100K. From these, we constructed multiple variants with different forms of illumination variation.

\begin{figure}[htb!]
  \centering
   \includegraphics[width=0.9\linewidth]{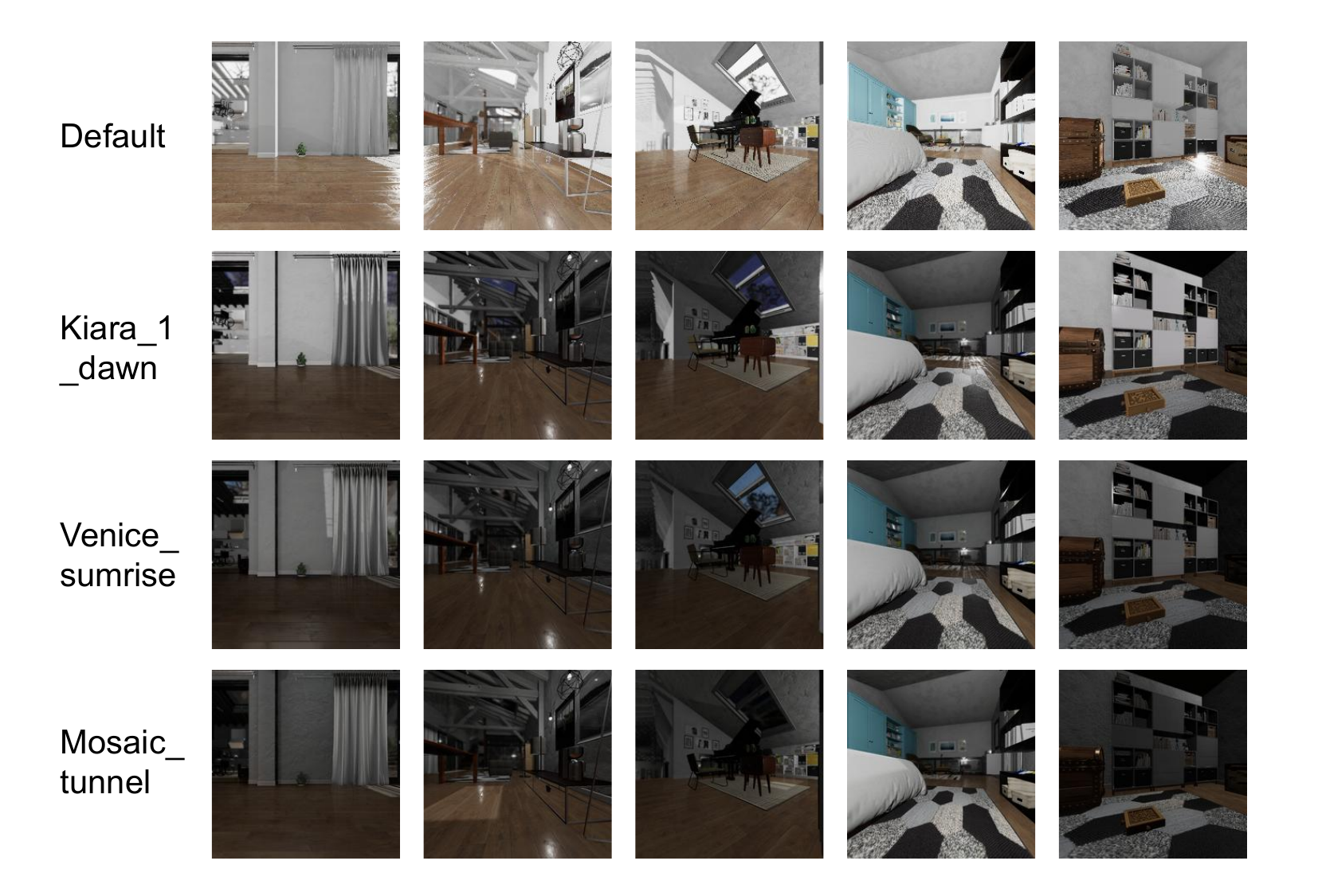}
   \caption{Examples of the House100K dataset captured at the same spatial locations under the default lighting environment and three skyboxes.}
   \label{fig:skybox}
\end{figure}

In the House100K dataset, images are captured under default lighting conditions in a controlled virtual environment using ThreeDWorld (TDW)~\citep{gan2020threedworld}, where a virtual agent traverses a predefined trajectory consisting of 102,197 steps. The\break \textbf{House100KLighting} variant introduces illumination variation by collecting ten images for each step, including one under default lighting and nine under diverse, realistic lighting conditions (called skyboxes; see Figure~\ref{fig:skybox}). Positive pairs for contrastive learning are defined based on thresholds on positional and rotational differences, following ESS-MB~\citep{zhu2024incorporating}.

ImageNet100K was created by randomly selecting 100 images from each of the 1,000 ImageNet categories~\citep{5206848}. 

\begin{figure}[ht!]
    \centering
    \includegraphics[width=0.7\linewidth, trim=0.8cm 0.6cm 1cm 0.6cm, clip]{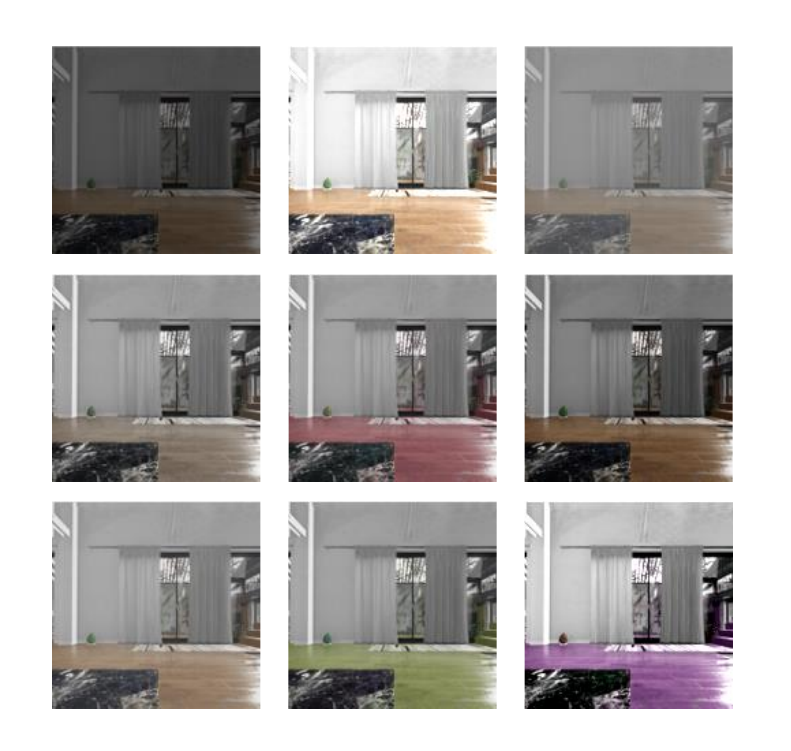}
    \caption{
    Nine image-level transformations used to approximate illumination
    variation, including adjustments to brightness, contrast, saturation,
    hue, and gamma. Numbers in parentheses denote the corresponding
    transformation parameters.
    }
    \label{fig:simplistic}
\end{figure}

To study settings without access to physically based rendering, we created augmented-lighting variants of both datasets, denoted as \textbf{House100K-AugLighting} and\break \textbf{ImageNet100K-AugLighting}, by applying a set of standard image-level transformations (i.e., brightness, saturation, contrast, hue, and gamma adjustments; see Figure~\ref{fig:simplistic}) to approximate illumination changes.

\subsection{Baseline and Metrics}
To evaluate the quality of the learned representations, we froze the backbone model and trained task-specific heads for multiple downstream tasks. The evaluation tasks and corresponding metrics are as follows:\\
\textbf{ImageNet classification}: Evaluated on the standard 1,000-class classification task, using approximately 1.28 million images for training and 50,000 images for validation.\\
\textbf{Exdark object and lighting classification}~\citep{Exdark}: Evaluated using around 12,000 indoor and outdoor images for both object classification (12 classes) and lighting classification (10 classes).\\
\textbf{PASCAL VOC detection}~\citep{Everingham10}: Evaluated on the combined PASCAL VOC 2007 and 2012 datasets, comprising 16,551 images across 20 object categories.

All experiments were repeated three times, and results are reported as averages to account for variability and ensure statistical reliability. 

\subsection{Dual-Head Lighting-aware Training Outperforms the Counterpart}
We conducted experiments to evaluate the \textbf{dual-head, lighting-aware training} approach (configuration $(1,0,7)$) in comparison with \textbf{standard contrastive training} $(0,0,8)$ and a \textbf{single-representation baseline} $(0,8,0)$, which is trained using both lighting and standard contrastive objectives. All models were pretrained on House100KLighting. Unless otherwise specified, we use $\alpha = 1$ to balance the content and light loss.

For clarity, we denote the representations used in downstream tasks as follows: the lighting representation corresponds to the features learned by the lighting head, the content representation corresponds to features learned by the standard contrastive head, and the combined representation is obtained by concatenating all components. In the dual-head setting, we evaluated downstream classification using each of these representations, whereas for standard and single-representation basedlines, only the combined representation was used.

\begin{table*}[ht!]
    \centering
    \begin{tabular}{l l l c c c }
    \toprule
        Repr. & \multicolumn{2}{c}{Pretext Task} & \multicolumn{3}{c}{Downstream ImgNet Accuracy}\\
        prop. & Con. loss & Light loss &Combined&Content&Light\\
        \hline
        (1,0,7) & $4.48\pm 0.02$ & $7.77\pm 0.02$ & $21.70\pm 0.38$ & $20.55\pm 0.63$ & $10.69\pm 0.35$\\
        (0,0,8) & $4.03\pm 0.00$ & N.A. &$20.23\pm 0.12$ &N.A.&N.A.\\
        (0,8,0) & $4.83\pm 0.00$ & $7.62\pm 0.00$ & $20.62\pm 0.33$&N.A.&N.A.\\
        (1,1,6) & $4.50\pm 0.01$ & $7.75\pm 0.00$ & $\textbf{21.85}\pm 0.35$&$\textbf{21.19}\pm 0.29$&$\textbf{16.24}\pm0.80$\\
        \hline
        (1,0,1) & $4.91\pm 0.10$ & $7.33\pm 0.11$ &$19.36\pm0.02$ &$14.94\pm 0.02$&$15.09\pm 0.52$\\ 
        (1,0,5) & $4.50\pm 0.00$ & $7.75\pm 0.00$ &$21.44\pm 0.53$&$20.03\pm 0.37$&$11.85\pm 0.38$\\
        (1,0,9) & $4.25\pm 0.23$ & $8.04\pm 0.28$ & $21.38\pm 0.38$&$20.47\pm0.70$&$10.70\pm 1.94$\\
    \bottomrule
    \end{tabular}
    \caption{\textbf{Comparison of dual-head training and other contrastive settings on the ImageNet classification task.} Repr. prop. is the proportion of the representation. (L:J:C) denotes the proportion of the dimension allocated to lighting, joint, and content representations, respectively. 'Content', 'Light', and "Combined" represent the content, light-sensitive and overall (light+content+joint) representations. Numbers after $\pm$ represent the standard error of the mean, rounded to a minimum of .01. }
    \label{tab:main}
\end{table*}

\begin{table*}[ht!]
    \centering
    \small
    \setlength{\tabcolsep}{3pt}
    \begin{tabular}{l|ccc|ccc|c}
    \toprule
        Repr. & &ExDark Obj.&&&ExDark Light&& \multicolumn{1}{c}{Pascal Det.}\\
        prop. &Comb.&Cont.&Light&Comb.& Cont.&Light&Comb. AP\\ 
        \hline
        (1,0,7) & $\textbf{28.9}\pm 0.2$ & $29.9\pm 0.1$ & $20.7\pm 0.4$ & $\textbf{52.8}\pm 0.2$ & $52.1\pm 0.3$& $48.6\pm 0.4 $&$\textbf{48.9}\pm 0.2$\\
        (0,0,8) &$28.7\pm 0.4$ &N.A.&N.A.&$50.1\pm 0.4$&N.A.&N.A.&$48.6 \pm 0.1$\\
    \bottomrule
    \end{tabular}
    \caption{\textbf{Comparison of dual-head training and standard contrastive learning on ExDark dataset and Detection task.} Comb., Cont., and Light denote combined, content-only, and lighting-only representations, respectively.}
    \label{tab:extra}
\end{table*}

As shown in Table~\ref{tab:main}, rows 1--4 indicate that the dual-head model $(1,0,7)$ outperforms both standard contrastive training and the single-representation baseline when using the combined representation. This suggests that explicitly separating illumination-aware and content features during training yields more robust representations by mitigating the structured uncertainty introduced by varying illumination. Furthermore, we observe even greater performance gains when using representations that include joint components, such as $(1,1,6)$, which preserve interactions between content and illumination factors. As shown in Table~\ref{tab:extra}, the advantage of the combined representation learned through dual-head training is consistently observed across both object classification and lighting condition recognition tasks.

We also observe that, when used individually, the illumination and content heads are less effective than the combined representation, confirming that both heads capture complementary information. On ImageNet (Table~\ref{tab:main}), the content representation dominates object classification, whereas on low-light datasets such as ExDark (Table~\ref{tab:extra}), the illumination representation contributes substantially, in some cases surpassing the content representation. Interestingly, the stronger performance of partially shared representations such as $(1,1,6)$ suggests that retaining limited interaction between semantic and illumination-related features may be beneficial for downstream recognition.

\begin{figure}[ht!]
\begin{center}
    \includegraphics[width=0.9\linewidth]{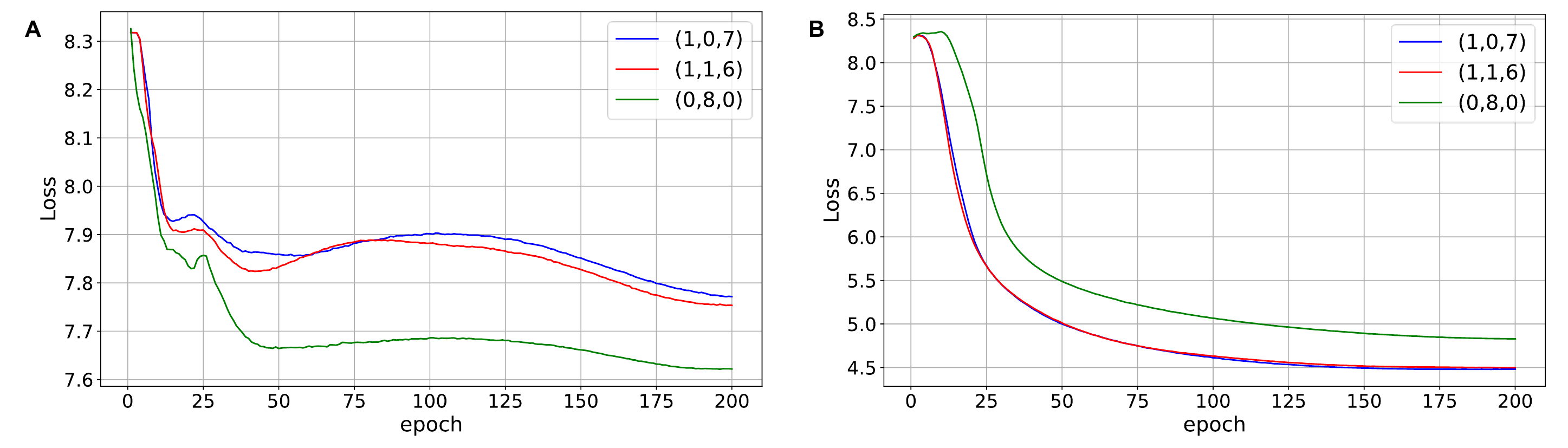}
\end{center}
\caption{Pre-training loss curves of lighting-aware approaches on the House100KLighting dataset: (A) lighting loss and (B) contrastive loss.} 
\label{fig:curve}
\end{figure}

\begin{figure}[ht!]
\begin{center}
    \includegraphics[width=0.9\linewidth]{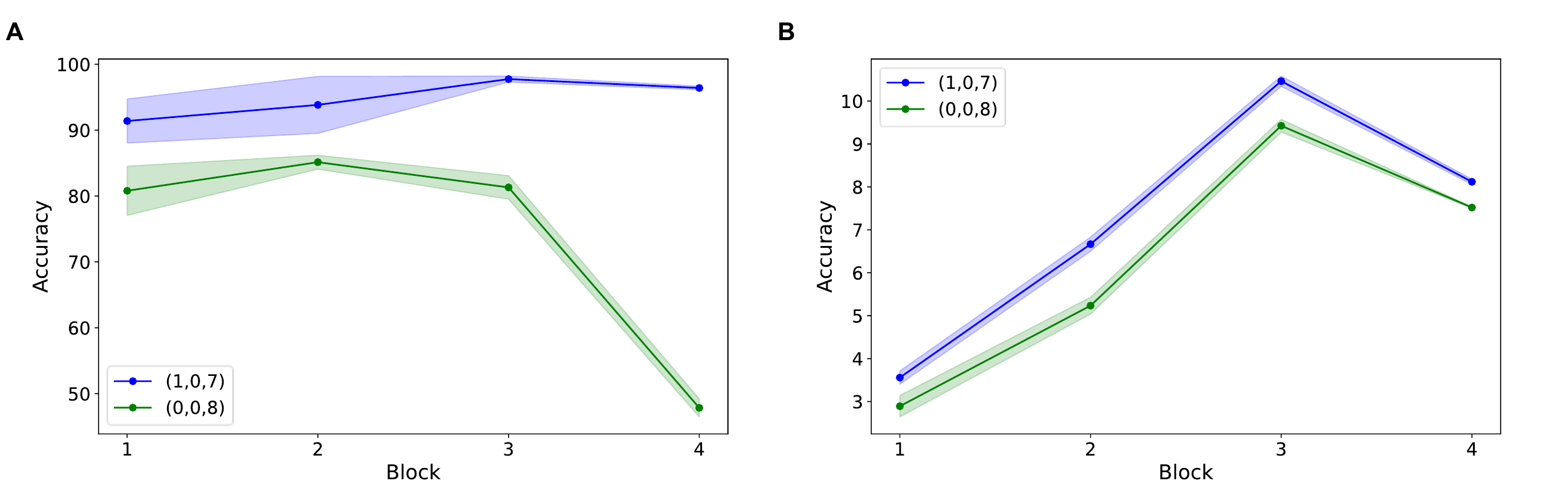}
\end{center}
\caption{Comparison of layer-wise internal embeddings learned by dual-head $(1,0,7)$ and standard contrastive training $(0,0,8)$. Accuracy of (A) lighting prediction task (B) small ImageNet classification task.} 
\label{fig:layer}
\end{figure}

Notably, as shown in Figure~\ref{fig:curve}, the lighting loss curves for the pretrained configurations $(1,0,7)$, $(1,1,6)$, and $(0,8,0)$ exhibit non-monotonic behavior. The rapid decrease in the first ten epochs likely reflects the model's ability to capture relatively simple illumination cues, whereas the subsequent fluctuations after epoch 10 suggest increasing influence of interaction between illumination and content factors. As training progresses, the gradients appear to be increasingly influenced by higher-level semantic features associated with object classification. In contrast, the single-representation approach shows a relatively stable lighting loss after epoch 50, likely because it conflates content and lighting information at that point.

\begin{table*}[ht!]
    \centering
    \scalebox{0.9}{
    \begin{tabular}{l l l l c c c}
    \toprule
        Pre. &Repr. & \multicolumn{2}{c}{Pretext Task} &\multicolumn{3}{c}{Downstream Small ImgNet Accuracy}\\
        Dataset& Prop. & Con. loss & Light loss &Combined& Content&Light\\
        \hline
        House100K&(1,0,7)&$4.42\pm 0.11$&$7.82\pm 0.13$&$\textbf{18.42}\pm 0.22$&$18.44\pm 0.19$&$7.95\pm 0.22$\\
        Lighting&(0,0,8)&$4.00\pm 0.00$&N.A.&$18.00 \pm 0.12$&N.A.&N.A.\\
        \hline
        \multirow{2}{*}{ImageNet100K}&(1,0,7)&$4.76\pm 0.03$&$7.98\pm 0.05$&$\textbf{21.42}\pm 0.25$&$20.58\pm 0.41$&$9.23\pm 0.24$\\
        &(0,0,8)&$4.42\pm 0.00$&N.A.&$20.93\pm 0.34$&N.A.&N.A.\\
    \bottomrule
    \end{tabular}
    }
    \caption{\textbf{Comparison of models pre-trained on various datasets with simplistic lighting conditions.} }
    \label{tab:aug}
\end{table*}

\begin{table*}[ht!]
    \centering
    \begin{tabular}{l cc cc}
    \toprule
        Repr. prop. & \multicolumn{2}{c}{Object Classification} & \multicolumn{2}{c}{Lighting Classification}\\
        \cmidrule(r){2-3} \cmidrule(r){4-5}
         & Loss & Acc. & Loss & Acc. \\
        \hline
        (1,0,7) & $0.47\pm 0.01$ & $\textbf{43.51}\pm 0.14$ & $0.64\pm 0.00$ & $74.85\pm 0.64$ \\
        (0,0,8) & $0.38\pm 0.00$ & $42.30\pm 0.09$ & N.A. & N.A. \\
    \bottomrule
    \end{tabular}
    \caption{\textbf{Results of dual-head training in the supervised training setting.}}
    \label{tab:supervise}
\end{table*}

\subsection{Layer-wise Analysis}
\label{sec:layer}
To examine how illumination and content information are distributed across network depth, we designed layer-wise probing tasks to evaluate classification performance at different stages of the trained model. Specifically, we tested whether illumination-aware representations propagate throughout the network or are primarily concentrated in later layers near the task-specific heads.

We constructed two probing tasks: (i) a small ImageNet classification task and (ii) a supervised illumination classification task. For the ImageNet task, we sampled 50 images per class from the training set and used the standard validation set for evaluation. For the illumination task, 100,000 images from House100KLighting with skybox labels were sampled and evenly split into training and testing sets. For both tasks, a linear classifier was attached to the output of each ResNet-50 block.

As shown in Figure~\ref{fig:layer}, the dual-head training approach consistently outperforms standard contrastive learning, particularly on the illumination classification task. All results show an improvement in the first three blocks but a slight decline in the final block. This decline is especially evident under standard contrastive training, where illumination classification accuracy in the final layer decreases substantially, by nearly half. We hypothesize that later layers become increasingly optimized for semantic discrimination, reducing sensitivity to illumination-related visual structure. This effect is more pronounced in standard contrastive learning, where illumination information is only indirectly introduced through augmentation, potentially limiting the preservation of illumination-related features in deeper layers.

\subsection{Generalization to Simple Image Operations}
We evaluated whether simple image transformations could approximate lighting variations. As shown in Table~\ref{tab:aug}, results on both House100K-AugLighting and ImageNet100K-AugLighting indicate that models trained with the dual-head approach achieve only marginal improvements over standard contrastive learning, with differences falling within the variability observed across repeated runs. Moreover, overall performance remains lower than that obtained using ray-traced illumination. 

These results suggest that simple image-level transformations provide only a limited approximation of realistic illumination variation. While operations such as brightness and contrast adjustment can introduce low-level appearance changes, physically based lighting effects involve substantially more complex interactions between geometry, materials, and illumination conditions.

\subsection{Generalization to Supervised Learning}
Although simple image transformations cannot fully replicate realistic illumination, they enable supervised training with both object and lighting labels. We therefore investigated the dual-head framework on ImageNet100K-AugLighting in a supervised learning setting. The output of the ResNet average pooling layer was divided into two parts, corresponding to illumination and content representations. The model is trained jointly on two tasks: object classification and illumination condition classification. The single-head baseline is trained only on the object classification task. In the dual-head setting, the illumination-aware representation was evaluated only on the lighting classification task, while the semantic representation was evaluated only on the object classification task. Consequently, each downstream task used only a subset of the full representation dimensions. In contrast, the single-head baseline used the full representation for object classification.

As shown in Table~\ref{tab:supervise}, dual-head training shows a slightly higher training loss but achieves improved test accuracy despite using reduced representation dimensionality for each task, indicating enhanced generalization through multi-factor supervision.

\subsection{Ablation Study}
We analyzed the effect of varying the allocation of representation capacity between lighting and content. As shown in Table~\ref{tab:main} (rows 6--7), slightly increasing or decreasing the dimensionality assigned to the lighting representation has minimal impact on downstream tasks. However, when the lighting head proportion is increased to half of the final output layer, specifically $(1,0,1)$, classification performance degrades significantly, even when using the combined representation from both heads.

These results suggest that illumination-aware supervision is beneficial when balanced with sufficient representation capacity for semantic discrimination, whereas excessive allocation toward illumination-related features may reduce overall recognition performance.

\begin{table*}[ht!]
    \centering
    \begin{tabular}{l l l c c c }
    \toprule
        $\alpha$ & \multicolumn{2}{c}{Pretext Task} & \multicolumn{3}{c}{Downstream ImgNet Accuracy}\\
         & Con. loss & Light loss &Combined&Content&Light\\
        \hline
        2 & $5.95\pm 0.02$ & $6.58\pm 0.00$ &$19.41\pm 0.30$ &$19.55\pm 0.09$&$7.54\pm 0.03$\\
        1 & $4.48\pm 0.02$ & $7.77\pm 0.02$ & $\textbf{21.70}\pm 0.38$ & $\textbf{20.55}\pm 0.63$ & $10.69\pm 0.35$\\
        0.5 & $4.02\pm 0.00$ & $8.32\pm 0.00$ & $21.02\pm 0.08$& $19.22\pm 0.23$& $\textbf{13.94}\pm 0.08$\\
    \bottomrule
    \end{tabular}
    \caption{\textbf{Ablation study on the weighting coefficient $\alpha$ between the contrastive and lighting-aware losses of the ImageNet classification task.}}
    \label{tab:weight}
\end{table*}

We further study the weighting coefficient $\alpha$ between the two loss terms and find that the model is relatively insensitive to moderate variations, with $\alpha = 1$ achieving the best performance compared to $\alpha = 0.5$ and $\alpha = 2$, as shown in Tabale~\ref{tab:weight}.

\subsection{Discussion}

Our results demonstrate that explicitly incorporating illumination-aware information improves model performance. However, our approach has several limitations. First, the internal propagation of illumination and content signals unclear. Investigating layer-wise dynamics or introducing intermediate supervision could provide deeper insight into how representations evolve throughout the network. Second, experiments are conducted on relatively small datasets for tractability; extending to larger and more diverse datasets is necessary to more rigorously assess generalization. Third, while simple image transformations provide an efficient approximation of illumination variation, they cannot fully capture complex physical lighting effects, limiting performance in realistic scenarios. 

Future work could extend this factor-aware learning framework to additional controllable factors, such as viewpoint, color or scale variation, to further improve robustness across environments. Another promising direction is to explore efficient methods for generating realistic training datasets, for example, using ICLight~\citep{zhang2025scaling} or other ControlNet-based~\cite{zhang2023adding} models to manipulate illumination or other factors while preserving scene semantics.

\section{Conclusions}

This study suggests that illumination need not be treated solely as a nuisance factor in representation learning. By explicitly incorporating lighting information into the training objective, the proposed framework learns representations that retain both semantic and illumination-dependent structure, leading to consistent improvements across downstream visual recognition tasks. Beyond the empirical gains, our results indicate that environmental variation can provide useful signals rather than merely serving as a source of variation to be suppressed.

More broadly, we view illumination as a representative example of a controllable factor that influences image formation. The ability to manipulate such factors in simulated environments creates new opportunities to study how representations respond to controlled environmental variation. We hope that this perspective encourages future work on factor-aware representation learning, where controllable aspects of the visual world are leveraged to improve robustness, generalization, and understanding of learned representations.


\acks{This material is based upon work supported in part by the National Science Foundation under Grant No. BCS-2216127. This work used cluster computers at the Pittsburgh Supercomputer Center through an allocation from the Advanced Cyberinfrastructure Coordination Ecosystem: Services \& Support (ACCESS) program, which is supported by NSF Grants Nos. 2138259, 2138286, 2138307, 2137603, and 2138296. }


\newpage

\appendix
\section{Experimental Configurations}
\textbf{Pretraining of contrastive learning:} All simulations were executed thrice on four NVIDIA RTX A6000 GPUs, with the 200 epoches and 256 batchsize. A memory bank size of 4096 and a 128-dimensional output are utilized. All models are optimized using the SGD optimizer with a cosine learning rate schedule. The temperature coefficient is set to 0.2, and the momentum update rate is 0.999. For ESS-MB models, a learning rate of 0.3 is used. The standard MoCo V2 uses a learning rate of 0.03. For models trained with different lighting conditions using ray tracing, we applied the following data augmentation strategies: random cropping with a scale range of (0.2, 1), random grayscale transformation with a probability of 0.2, color jittering with a probability of 0.1, Gaussian blur, and random horizontal flipping. For the models trained with simple image transformations, we apply the same augmentation methods except color jittering to avoid redundant operations.\\
\textbf{Downstream Evaluation:} For ImageNet classification tasks pre-trained with ray-traced lighting conditions, two out of three runs were conducted on a high-performance computing platform with four shared V100 GPUs in the same trail, while all other training was performed on four NVIDIA RTX A6000 GPUs.
For all the linear classification tasks, including Exdark object and lighting classification, ImageNet classification and House100KLighting lighting classification, we add a linear layer at the top of the frozen ResNet-50 backbone. All of them are trained with 50 epoches and SGD optimizer. For ImageNet classification, we started with a learning rate of 30.0, as in MoCoV2. A learning rate of 0.3 was used other datasets, to ensure faster convergence.

For the PASCAL detection task, we used Faster R-CNN as the detector and fine-tuned it with a pre-trained ResNet-50 backbone. The model was trained for 22,000 iterations with a learning rate of 0.2.\\
\textbf{Supervised training: }Supervised learning was conducted using the ResNet-50 architecture. The dual-head output was divided into lighting and content features in a 1:7 ratio. Training was performed using the SGD optimizer with a cosine learning rate schedule starting at 0.03. 

\section{Lighting Supervision Experiment}
We further investigate an alternative training strategy for the lighting-aware branch by replacing the contrastive objective with a supervised regression-style formulation. In this variant, the lighting representation is trained using explicit supervision derived from skybox labels, which provide direct information about illumination conditions in the synthetic environment. Specifically, while the content branch is still optimized using standard contrastive learning, the lighting branch is trained to predict the corresponding skybox-based lighting identity. This design allows us to assess whether explicit supervision can serve as a stronger or more stable learning signal compared to contrastive objectives for illumination modeling, without modifying the overall architecture or representation dimensionality.

\begin{table*}[ht!]
    \centering
    \begin{tabular}{l l l c c c }
    \toprule
        Lighting & \multicolumn{2}{c}{Pretext Task} & \multicolumn{3}{c}{Downstream ImgNet Accuracy}\\
         traain mode& Con. loss & Light loss &Combined&Content&Light\\
        \hline
        Contrastive & $4.48\pm 0.02$ & $7.77\pm 0.02$ & $21.70\pm 0.38$ & $20.55\pm 0.63$ & $10.69\pm 0.35$\\
        Supervised & $4.39\pm 0.00$ & $0.06\pm 0.00$ & $21.77\pm 0.07$& $21.51\pm 0.05$& $8.68\pm 0.02$\\ 
    \bottomrule
    \end{tabular}
    \caption{\textbf{Comparison between contrastive and supervised learning for the lighting-aware branch.}}
    \label{tab:sup_light}
\end{table*}

Table~\ref{tab:sup_light} compares the fully contrastive formulation with the mixed setting where only the lighting branch is supervised on the downstream ImageNet classification task. Introducing supervised learning for the lighting branch yields comparable performance on the combined representation, while leading to a more pronounced improvement in the content branch and a notable degradation in the standalone lighting representation.

This suggests that explicit supervision encourages the model to prioritize features that are more directly aligned with the lighting classification objective, rather than learning a structured and transferable illumination-aware embedding space. One possible explanation is that supervised learning constrains the lighting branch to focus on discriminative signals specific to the provided labels, which may not fully capture the continuous and compositional nature of illumination variation. As a result, while the model allocates less representational capacity to modeling illumination explicitly, it benefits content feature learning by reducing competition for shared representational resources. However, completely discarding a structured lighting representation is also suboptimal, as some degree of illumination-aware modeling appears necessary to support robust generalization, particularly when combining representations from both branches.

\vskip 0.2in
\bibliography{sample}

\end{document}